\documentclass[a4wide]{article} 

 \usepackage{mathptmx}      
\usepackage{graphicx}
\usepackage{natbib} 
\usepackage{url}
\usepackage{gb4e} 

\newcommand\flex{ \textsc{(f)}}  
\newcommand\rig{ \textsc{(r)}} 
\renewcommand\phi\varphi 

\newcommand\fl{\mathbin{\rightarrow}}
\newcommand\ttt{\mathsf{t}}
\newcommand\eee{\mathsf{e}}
\newcommand\vvv{\mathsf{v}}
\newcommand\xxx{\mathsf{x}}
\newcommand{\ltyn}{\ensuremath{\Lambda\!\mathsf{Ty}_n}}
\newcommand\systF{\mathsf{F}}
\newcommand\Land{\&^\Pi}
\newcommand\ma[1]{\textit{``#1''}}
\newcommand\lpl{\mathtt{lpl}}
\newcommand\assi{\mathop{\mathtt{assi}}}
\newcommand\atratres{\mathop{\mathtt{atra3}}}
\newcommand\furou{\mathop{\mathtt{furou}}}
\newcommand\ilg{\mathop{\mathtt{ilg}}}
\newcommand\sig{\mathop{\mathtt{sig}}}

\newcommand\keywords[1]{\noindent\textbf{Keywords:}  #1}

\begin{document}

\title{Deverbal semantics\\ and the Montagovian generative lexicon $\ltyn$
}


\author{Livy Real (Universitade Federal do Paran\'a, Curitiba) \and Christian Retor\'e (LaBRI, Universit\'e de  Bordeaux)}


%

\maketitle

\begin{abstract}
We propose a lexical account of action nominals, in particular of deverbal nominalisations, whose meaning is related to the event expressed by their base verb. The literature about nominalisations often assumes that the semantics of the base verb completely defines the structure of action nominals. We argue that the information in the base verb is not sufficient to completely determine the semantics of action nominals. We exhibit some data from different languages,  especially from Romance language, which show that nominalisations focus on some aspects of the verb semantics. The selected aspects, however, seem to be idiosyncratic and do not automatically result from the internal structure of the verb nor from its interaction with the morphological suffix. We therefore propose a partially lexicalist approach view of deverbal nouns. It is made precise and computable by using the Montagovian Generative Lexicon, a type theoretical framework introduced by Bassac, Mery and Retor\'e in this journal in 2010. This extension of  Montague semantics with a richer type system easily incorporates lexical phenomena 
like the semantics of action nominals in particular deverbals, including their polysemy and (in)felicitous copredications. 

\end{abstract}

\section{Introduction}
\label{intro}

This paper rather deals with the linguistic side of word meaning. It  is about  the semantics of nominalisations and their integration into a computational  framework for compositional semantics. 
We argue for a rather lexicalist view of deverbal nominalisations. We  propose  a lexical characterisation, formalised in the Montagovian Generative Lexicon framework, to polysemous nominalisations. 
We are especially concerned with the polysemy between processual and resultative readings, found in many different languages. Here, we mainly consider some romance languages (French, Italian,  Brazilian   Portuguese),  English, German but apparently our observations generalise to other  germanic and romance languages. We mainly look at nouns that consists in a verb and suffix, as \textit{construction} (\textit{construct} + \textit{tion}), and we leave aside nouns as \textit{travel} and \textit{run}, which do not use any suffix and thus can be expected to be quite different ---- furthermore Romance languages that we focus on do not have any deverbal identical to the verb.  In this paper,
the main focus is on eventive nominalisations, or,  as defined by \citet{melloni}, on action nominals:

\begin{quotation} \it 
Specifically, action nominals are headed by suffixes conventionally named as `transpositional' in the linguistic literature (cf. \citet{Bea95} for such definition), because they simply transpose the verbal meaning into a semantically equivalent lexeme of category N. In effect, according to \cite[p.178]{Com76}, action nominals are ``nouns derived from verbs (verbal nouns) with the general meaning of an action or process''. \citep[p.8]{melloni}
\end{quotation}

Generally, action nominals denote, beyond the event/action, the result or the resultative state of this action,
and we shall focus on such issues: 

\begin{exe}
\ex \label{8months} The construction took eight months thanks to our volunteers and staff.  (event)
\ex \label{style} As you can see the majority of the construction is of traditional style. (result)
\end{exe}

In sentence \ref{style}, \textit{construction} has a resultative interpretation forced by \textit{style}, meanwhile in \ref{8months},  \textit{construction} has an eventive reading as it is the argument of \textit{took eight  months}.

In contrast with most linguists who paid attention to this field, we believe that the syntatic-semantical features of nominalisations, depend heavily on contexts and  cannot be completely predictable from the semantics of their base verb only. We also observed that action nominals may have much more diverse  relations with their base verbs, than event or result (the location is possible): approaches that leave out other senses are mistaken.  We present data, mainly from Romance languages and from time to time from  Germanic languages: they all  show some lexical arbitrariness. This is a good reason for   a formalisation of the behaviour of those nominals with  the Montagovian Generative Lexicon, whose architecture is more word-driven than 
type-driven. 

We first present an overview about nominalisation studies. Thereafter, we briefly outline the current beliefs concerning the nature of deverbal nominals, summarising  quickly  the discussions of \citet{grim} and \citep{jezek}. Then, we present some data that cannot be captured by the generalisations already that other authors proposed. Finally, we recall the Montagovian Generative Lexicon of \citet{BMRjolli} and present in this framework, a computational account of the semantic behaviour of nominalisations in contexts.

\section{Nominalisations}
\label{sec:1}

Nominalisations are nouns derived from other syntactical categories, especially deverbals that derive from verbs. 
%
%
The study of nominalisations has been receiving  great attention from linguists in the last decades.
The morpho-syntactical features of nominalisations have been studied since at least 1970 (see e.g. \citep{Cho70}) and their syntactical-semantical features for at least two decades (e.g. \citep{grim}). Recently, the characterisation of deverbal nouns also has been considered pragmatically and ontologically \citep{Ham09,bra11}.
The interest for nominalisations on modern linguistics increased with \citet{Cho70}
and the subsequent \emph{Lexicalist Hypothesis} issued from his view. Chomsky, considering English data, argued that nominalisations are, in deep structure, nouns rather than transformations from verbs. Some contemporary works on the  computational semantics of nominals also try to relate base verbs  and their nominalisations, but, according to \citet{Gur06}, many parsers can analyse \ref{destroy} , but not \ref{destruction}: 

\begin{exe}
\ex 
\begin{xlist}
\ex Alexander destroyed the city in 332 BC. \label{destroy}
\ex Alexander’s destruction of the city happened in 332 BC. \label{destruction} 
\end{xlist}
\end{exe}

Indeed, it is not easy  to compute  of the arguments of the verb (\textit{destroy}) when a nominalisation (\textit{destruction}) introduces the event. How the arguments of the base verb are mapped to the nominal structure led many scholars to propose some sort of argumental structure to these nominals issued from the argumental (or eventive) structure of the base verb, but as the forthcoming examples will show, they are doubtful.

Different linguistic questions appear when looking at nominalisations. Some of them are general questions, like a proper account of  word formation \citep{Jac75},  or like the  morpho-syntactical parallel between pairs of sentences with nominalisations and  one with the corresponding verb \citep{Cho70}. 
More recently, more specific questions are addressed like  the relation between verb argumental structure and the  argument structure of the corresponding nominalisation \citep{grim}. 
A computational treatment of 
 pairs of sentences involving verbs and nominalisation is now expected  \citep{Gur06}. 

Most of these questions  already received at least a partial treatment by linguists. 
However some questions remain unsolved like that of 
the polysemous behaviour of nominalisations in relation to multiple predications. 
We shall study them within the  Montagovian Generative Lexicon which properly computes the meaning of nominalisations which are ambiguous between different aspects, commonly the result and the process,
but many other senses are possible.

%

In the next section, we discuss two different proposals about  nominalisations, before presenting our proposal. 

 \section{Nominalisations in the  literature}\label{sec:2}

Let us present two different discussions about the nature of action nominals: the pioneering work by \citet{grim} and more recent description within the  Generative Lexicon framework, as \citet{Jacquey2006tal,asher-webofwords,jezek}. Up to now, the first work remains  a major reference on nominalisations, especially in Chomskyan linguistics. The Generative Lexicon considers nominalisations as a particular case of nominals with a complex type (dot-types). In this setting, authors mainly discuss the behaviour of nominalisations in co-predication contexts. The literature from the last five years, including \citet{jezek}, even handle more complicated phenomena in  the Generative Lexicon framework, so we shall also pay attention to these recent approaches.


\subsection{Grimshaw's seminal work}

The study by \citet{grim} is probably the most famous proposal dealing with the internal structure of those nominals. We shall discuss two main points from her work: 
\begin{itemize} 
\item 
the  inheritance of the argumental structure of the nominal from that of the base verb and 
\item 
the pluralisation of action nominals that denote events. 
\end{itemize} 

Grimshaw already noticed the polysemy between the event reading and the result reading. She tells apart  eventive nouns --- ``complex events'' --- from ``result nouns'', the latter ones being part of a bigger class called ``simple events''.  Grimshaw, based on English data, assumes that complex events are nominalisations that preserve the entire argumental structure of their base verb and that result nouns do not inherit any argumental structure from their base verbs. \citet{grim} also notices that complex events cannot be pluralised, as opposed to result nouns. From these observations, she claims that complex events act like non-count nouns and that result nouns act like count nouns. However, we think that this assertion does not always hold. 
Firstly, let us take a look at some examples of pluralised complex events before discussing the saturation of arguments.

%
%
%
%
%
%
%


\begin{exe}
\ex The several \textbf{destructions} of the Temple, and all their sufferings and \textbf{dispersions}, continued most wonderfully and identically the same down to the destruction of the Temple? \footnote{Chambers, John David. Lights before the Sacrament: an argument, scriptural, historical, and legal in a letter to a member of convocation, London, 1866.}\\
%
\ex The \textbf{translations} took many hours of hard, slogging work, often with material which, because of its archaic and technical nature, was extremely difficult.\footnote{\url{http://seinenkai.com/articles/noble/noble-shorin1.html}}\\
%
%
%
%
\ex Les fr\'equentes \textbf{destructions} des quartiers populaires (French - \citep{Jas06}).
\trans `The frequent destructions of popular quarters' 

%
%
%


\ex Saiba como acontecem as \textbf{contagens} de votos para eleger vereadores e prefeitos.\footnote{\url{http://japerionline.com.br/japeri/}\linebreak\url{saiba-como-acontecem-as-contagens-de-votos-para-eleger-vereadores-e-prefeitos/}} (Brazilian Portuguese)
\trans `Learn how the vote counting for city councillors and mayors works'
 \end{exe}

As the examples above show, the regularities noticed by Grimshaw do not always apply, even in English. Grimshaw's conclusions about pluralised eventive nouns do neither hold in other languages, as established in recent literature:  Russian\citep{PA07}, Czech \citep{Pro06}, Japanese \citet{Myi99},  Portuguese \citet{SB07}, Romanian \citet{AIS2010nom}, German and Dutch \citet{Van91}.

Let us consider the internal argument structure proposed by \citet{grim}. She said that the differences between deverbal nominalisations designing a complex event and nominalisations that are a result noun come from the fact that the former inherit the argumental structure of their base verb  while the latter  do not. Let us remember Grimshaw's analysis on the following examples: 

\begin{exe}
\ex \label{hours} 
\begin{xlist} 
\ex {[Examination of the students] will take several hours.} \label{examstudents}
\ex {* [Examination] will take several hours.} \label{examhours}
\end{xlist}
\ex  \label{paper}
\begin{xlist}
\ex {* [The examination of the students] was printed on pink paper.} \label{examstudentspaper}
\ex    {[The examination] was printed on pink paper.} \label{exampaper} 
\end{xlist}
\end{exe}

In (\ref{examstudents}), \textit{examination} is a complex event and needs to be saturated by its arguments, however, in  \ref{paper},  \textit{examination} is a result noun, a simple event, and does not admit any argument: in particular it does not inherit any argumental structure from its base verb.
Many authors disagree on the necessity of postulating a direct or automatical inheritance of arguments from the base verb  by the nominalisations: \citet{Pic91} (Catalan), \citet{Oli06} (Brazilian Portuguese), \citet{Hey08} (English). Many languages contain examples with complex events that appear without any argument and examples with result nouns that admit arguments.

\begin{exe}
\ex  \label{pelo} 
\begin{xlist} 
\ex  result \citep{SB07}
\ex A an\'alise do texto pelo aluno enriqueceu o conhecimento dos colegas.
\trans `The analysis of the text by the student enriched the knowledge of the colleagues.'
\end{xlist}
\ex \label{dades} 
\begin{xlist}
\ex result - \citep{Pic91} 
\ex La discussi\'o de les dades es va publicar a la revista. 
\trans `The discussion of the data was published in the journal.'
\end{xlist}
\ex \label{priscian} 
\begin{xlist} 
\ex result and event \citep{melloni}
\ex La tua traduzione del testo di Prisciano, che \`e stata pi\`u volte corretta$_{[event]}$, \`e stata messa sulla scrivania$_{[result]}$.
\trans `Your translation of Priscian's text, which has been revised many times, was placed on the desk.'
\end{xlist} 
\end{exe}

In \ref{pelo}, in Portuguese, the arguments, \textit{do texto} (theme) and \textit{pelo aluno} (topic), of the nominalisation (\textit{an\'alise}) are present and the sentence is still felicitous. In Catalan \ref{dades}, also the sentence is felicitous keeping the resultative reading of \textit{discussi\'o} and the presence of its arguments. In \ref{priscian}, \textit{traduzione} has both readings (resultative and eventive one), even with the internal argument (\textit{del testo di Prisciano}) present.

This discussion shows that the behaviour of deverbal nominalisations can not be completely inferred from verbs and that similar deverbals from different related  languages may behave differently. 

\subsection{Generative Lexicon}
%
%
%
%
%

The behaviour of action nominals has been widely discussed in the Generative Lexicon literature \citep{Asher:93,AD05,Jacquey2006tal,melloni,jezek}, especially for finding out  the nature of the polysemy of action nominals on a par with other polysemous nouns, like \ma{ book} or \ma{newspaper}. Many scholars have been arguing that, at least in co-predication contexts, action nominals behave differently from other polysemous nouns: co-predication over different facets is usually infelicitous,  while it might be felicitous for other polysemous nouns. 

\begin{exe}
 \ex {The book is heavy but interesting.} \label{heavy} 
\ex[*] {The construction is mainly of traditional style and took eight months.} \label{tradi} 
\end{exe}

The example \ref{tradi} is a typical case of prohibited co-predication between these two different senses, process and result. On the other hand, in the example \ref{heavy},  the different meanings of \textit{book} (informational content and physical object, let us say) are felicitously coordinated. Apparently, this pattern of co-predication also holds in Brazilian Portuguese, French and Italian, as (\ref{cezanne}) below:
 
\begin{exe}
 \ex[*] {Les reproductions de C\'ezanne sont accroch\'ees$_{[result]}$ au mur et ont \'et\'e effectu\'ees$_{[event]}$ il y a peu. \citep{Jacquey2006tal}} \label{cezanne} 
\trans `The reproductions of C\'ezanne are hung on the wall and made not long ago.''
\end{exe}
%

As (\ref{cezanne}) shows, many nominalisations do not accept the co-predication between their different readings. But, in certain contexts, as (\ref{palladio}) and (\ref{assemblee}) below, the co-predication becomes felicitous. According to \citet{jezek}, there are syntactical constraints that allow the co-predication between these two different meanings of nominalisations: 

\begin{exe} 
\ex\label{syntacticalconstraints} 
\begin{xlist} 
\ex\label{clause} 
Split co-predication between main clause and subordinate clause; 
\ex\label{temporal} temporal disjunction between the two predications;
\ex\label{internal} omission of the internal argument.
\end{xlist} 
\end{exe} 

These constraints are exemplified below:  

\begin{exe}
\ex \label{palladio} La costruzione, che si protrasse fino al XVII secolo, rimane un\'importante testimonianza della geniale tematica del Palladio. \citep{jezek} 
\trans `The construction, which continued$_{[event]}$ till the XVII century, represents$_{[result]}$ an important evidence of Palladio’s ingenious artwork.'
\ex \label{assemblee} Les reproductions des s\'eances publiques sont effectu\'ees$_{[event]}$ conform\'ement aux r\`egles prescrites par l'Assembl\'ee, puis directement envoy\'ees$_{[result]}$ aux imprimeurs. \citep{Jacquey2006tal} 
\trans `The reproductions of the public meetings are conducted in accordance with rules prescribed by the Assembly, then sent directly to printers.'
\end{exe}





In \ref{palladio} and \ref{assemblee}, the constraints of (\ref{syntacticalconstraints}) 
seem to apply to the coordination between two different meanings of a single action nominal. 
Nevertheless, these constraints do not guarantee the felicity of copredications, as the following examples show:  

%

\begin{exe}
\ex[*]{La signature, qui est illisible$_{[result]}$, a pris$_{[event]}$ trois months.} \label{months} 
\trans ?“The signing/signature, that is illegible, lasted for three months.
\ex[*] {The examination, that lasted one whole day$_{[event]}$, was printed$_{[physic\_object]}$ in pink paper.} \label{whole} 
\ex[*]{A fritura, que sujou a cozinha ontem$_{[event]}$, est\'a muito boa$_{[result]}$.} \label{fritura} 
\trans `The fried food/frying, which soiled the kitchen yesterday, is very good.'
\end{exe}

Although the examples \ref{months}, \ref{whole}, and \ref{fritura} do satisfy  the constraints of  \citet{jezek},   they are not felicitous co-predications. We believe that the felicity of co-predication with action nominals is not only a matter of general rules. Indeed, the supposedly universal constraints in (\ref{syntacticalconstraints}) 
 does not seem to  work for every language and every action nominal. 
 
Conversely, there do exist copredications  that do not follow the constraints (\ref{syntacticalconstraints})   and yet  are clearly felicitous: 

\begin{exe}
\ex A fritura est\'a muito boa$_{[result]}$, ainda que tenha sujado a cozinha$_{[event]}$. \label{ainda} 
\trans `The fried food/frying is very good, although it has soiled the kitchen.'
%
%
\ex 1514 \"Uberreichte er Louis XII die schwierige \"Ubersetzung von Texten des Thukydides. \citep{bra11} \label{1514} 
\trans  ‘In 1514 he gave Louis XII the difficult translation of texts by Thucydides.’
\ex A constru\c{c}\~ao da cozinha est\'a \'otima$_{[result]}$, ainda que tenha demorado tr\^es dias$_{[event]}$. \label{ainda1} 
\trans `The construction of the kitchen is great, although it took three days.'
\end{exe}

Example (\ref{ainda}) shows that the co-predication between the eventive and the resultative senses of \textit{fritura} is possible, even if (\ref{fritura})  is not acceptable. Examples (\ref{1514}) and (\ref{ainda1})  show that even when an internal argument is  present, the co-predication can be felicitous.


As the data above show, the semantics of the base verb  and the syntactical configuration do not determine the behaviour of  action nominals.  For instance, many authors \citep{Cru04,asher-webofwords,bra11} have already noticed that discourse may affect the felicity of co-predication between (complex or simple) nominals.

\begin{exe}
 \ex
\begin{xlist}
\ex The city has 500 000 inhabitants and outlawed smoking in bars last year. \citep{asher-webofwords} 
\ex[?] {The city outlawed smoking in bars last year and has 500 000 inhabitants. \citep{asher-webofwords}} 
\end{xlist}
\ex
\begin{xlist}
\ex \label{tn} The newspaper was founded in 1878 and is still typed in Sutterin. \citep{bra11}
\ex \label{tn1} ?The newspaper was founded in 1878 and is printed in Frankfurt. \citep{bra11} 
\end{xlist}
\ex 
\begin{xlist}
 \ex \label{bar} Barcelona a organis\'e les jeux olympiques et gagn\'e quatre ligues des champions. 
\trans `Barcelona hosted the Olympic games and won four Champions Leagues.''
\ex[?] {Barcelona est la capitale de la Catalogne et a gagn\'e quatre ligues des champions.} \label{bar1}
\trans `Barcelona is the capital of Catalonia and our Champions Leagues.' 
\end{xlist}
\end{exe}

We can see that the specific context make the first sentence of each pair felicitous, while the second whose discursive context differs, is hardly acceptable . 
We are not yet able to account for such discursive or pragmatics factors 
that bias the standard semantic behaviour of deverbals. Nevertheless,  given our previous work 
on the Montagovian Geneative lexicon we think we can provide a formal and computational 
description of  of action nominals and their (in)felicitous co-predications as far as semantics is concern. 

%
%
%
%
%

 \section{Deverbals in the Montagovian Generative Lexicon}\label{sec:3}

Here we outline our proposal and show how it integrates into our computational setting which is partly implemented in Grail  \citep{moot10grail} --- more details on the logical framework in which this formal and computational modelling takes place can be found in some previous papers of us,  see for instance \citep{BMRjolli,Retore2012rlv,Retore2013taln}.
 
 Assuming a compositional semantical framework, we know that the base verb plays an important role on the formation of these nominals since most action nominals denote, among other senses, the same events or processes as the ones denoted by their base verb.  
Nevertheless  we also claim that the behaviour of action nominals cannot be completely inferred from the corresponding base verb.  
Indeed, we just  saw that no account of nominalisation in the literature is able to correctly predict all  and only the senses of a nominalisation.  For instance, nominalisations do not automatically inherit their arguments from the ones of the base verb (some are optional complements of the verb, for instance a location related to the event) 
and, contrary to  what  \citet{grim} said, many eventive nominals can be pluralised.  
 
 Many studies, including \citep{Jacquey2006tal} and \citep{jezek}, focus on the polysemy between eventive and resultative readings,  but  a nominalisation may refer to many more senses!  
 In German, following \cite[p.34]{bra11}, \textit{-ung} has eight different senses (event, result state, abstract result, result object, means, agent, collective, location). In Portuguese, considering the NOMLex-BR data base from \citep{Pai12}, \textit{ -\c{c}\~ao} has seven of those meanings, excluding the agent sense. In French, following the description from  the TLFi (Tr\'esor Informatis\'e de la Langue Fran\c{c}aise, \citep{Pierrel2006tlfi}), -\textit{age} could assume nine different senses (event, concrete result, abstract result, result object, means, agent, location, pejorative action). 

A diachronic study of nominalisation may provide good hints about the nominalisation process and the resulting senses. Some action nominals do not exhibit the eventive reading anymore, although they used to: this sense was lost at some point of language  evolution. In Portuguese, for example, \textit{estacionamento} (\textit{estacionar+mento}, \textit{parking}) only stands for a locative reading and never denote the action of parking, \textit{assadura}(\textit{assado+ura},\textit{roasting}) only stands for the resulting state, and so on.
 
Since an action nominal comes from a verb, the semantics of such a noun must be somehow related to the corresponding base verb, and it is.  But the semantical relation between the nominal and its verbal  base is not automatic, the senses of the deverbal do require some idiosyncratic information in addition to the suffix: 
nominalisations select an arbitrary specific part of the event represented by the verbal base.  This might be one or more senses from the event itself, its results (of all kinds), its location, its agent, and so on) but we cannot know in advance what kind of relation the nominal will establish with its base verb. 
For instance, if we consider the French nominalisation suffix  \textit{`-age'} 
it is hard to find a rule that can predict the sense(s) of \textit{dorage} (`browning'), \textit{maquillage} (`makeup'), \textit{t\'emoignage}  (`testimony'), \textit{garage} (`garage') \textit{p\^aturage} (`meadow').   They all stand in a different relation to their base verb: \textit{dorage} is the event itself, \textit{maquillage} the substance used, \textit{t\'emoignage} the result, \textit{p\^aturage} and \textit{garage} the location --- for the penultimate, the event reading is alsmot lost, and for the last one the event meaning is totally lost.

There are many  examples that show that in many languages the specific lexical-semantical content of a nominalisation is idiosyncratic. 
The meaning of a nominalisation comes from the history of a particular language and from the interaction between all the concurrent nominalisation suffixes available in the language being considered. In Portuguese, for instance, there can be three competing suffixes, for instance there exists: 
\begin{exe} 
\ex \label{les3armar} 
\begin{xlist}
\ex 
 \textit{armadura} (`armor'),
\ex  \textit{armamento} (`weapons') 
\ex \textit{arma\c{c}\~ao} (`preparation') --- that has other meanings like the event itself. 
\end{xlist} 
 \end{exe} 

They all  derive from \textit{armar} (`to arm'), which is polysemous between its original meaning, `to arm', and another derived meaning, `to set'. When \textit{arma\c{c}\~ao} denotes the event itself, it can 
only refer to `setting' and not to 
`arming'; thus, \textit{arma\c{c}\~ao} behaves differently from other words formed from \textit{armar} that select only the `arming' sense. 

Each nominal has been specialised and the semantics of the forming suffix does not firmly establish  the lexical relation held between the deverbal noun and the original verb. 
The suffixes \textit{-ura}, \textit{-mento} and \textit{-\c{c}\~ao} form action nominals 
that could be polysemous between eventive and resultative readings.  It happens that some nominalisations just reft to the event itself (\ref{feitura}), 
some of them can only refer to the result of the event (\ref{curvatura}), 
others are indeed polysemous between these two frequent senses
(\ref{destruicao})
and many others establish rarer relations with the main verb 
(\ref{estacionamento}).  

\begin{exe}
\ex \label{feitura} 
\begin{xlist}
\ex 
\textit{feitura} (`making'), 
\ex
\label{descobrimento} 
\textit{descobrimento} (`discovery', the resultative sense of `discovery' is denoted by \textit{descoberta}), 
\ex 
\label{giracao} 
\textit{gira\c{c}\~ao} (`gyration'), 
\end{xlist} 
\ex 
\label{curvatura} 
\begin{xlist} 
\ex 
\textit{curvatura} (`curvature'), 
\ex 
\label{comportamento} 
\textit{comportamento} (`behavior'), 
\ex
\label{legislacao}
\textit{legisla\c{c}\~ao} (`legislation'), 
\end{xlist} 
\ex \label{destruicao}
\begin{xlist}
\ex 
\textit{destrui\c{c}\~ao}(`destruction'), 
\ex 
\label{assinatura} 
\textit{assinatura} (`signature'/`signing'), 
\ex 
\label{desenvolvimento} 
\textit{desenvolvimento} (`development'),
\end{xlist}
\ex 
\label{estacionamento} 
\begin{xlist} 
\ex 
\textit{estacionamento} (`parking') denotes where to park (\textit{estacionar}), 
\ex 
\label{abotoadura} 
\textit{abotoadura} (`cufflink') a particular sort button to button the sleeves of a shirt  (\textit{abotoar}), 
\ex
\label{injecao} 
\textit{inje\c{c}\~ao} (`injection') what can be injected (\textit{injetar}). 
\end{xlist} 
\end{exe} 

Considering the lexical idiosyncrasies arising  in many different languages, 
we claim that the relation that action nominals establish with their base verb is partly arbitrary. 
Hence any uniform modelling of nominalisations would miss some senses, 
and we rather propose to specify in the lexicon the senses and their incompatibility  in accordance 
with the data and with the fact that speakers and electronic dictionaries must learn these types of  information. 
Nevertheless once this lexical and language specific information is known we propose
in the Montagovian generative lexicon a fully  automated 
analysis of the possible senses of sentences involving nominalisations 
which properly accounts for felicitous and infelicitous copredications. 
So our approach can be said to be partially lexicalist. 

\subsection{The Montagovian Generative Lexicon}

The standard compositional analysis of the semantics of a sentence consists in mapping inductively 
the (preferably binary) parse tree $t_s$ of a sentence $s$ 
to a logical formula  $[\![s]\!]$ which depicts its meaning. The lexicon provides each leaf of $t_s$,  that is 
a word $w_i$,  
with its semantics that is a $\lambda$-term $[\![w_i]\!]$ over the base types $\ttt$ (propositions) and $\eee$ (individuals).  
By structural induction on $t_s$,  we obtain  a  $\lambda$-term  $[s]{:}\ttt$ corresponding to $t_s$. 
Its normal form, that is a formula of higher order logic, is $[\![s]\!]{:}\ttt$, the meaning of $s$.
This standard process which implements Frege's compositionality principle is at the heart of Montague semantics. This computational and compositional view of semantics  relies on Church's representation of formulae as simply typed $\lambda$-terms, using the typed constants of figure \ref{church} --- see e.g  \cite[Chapter 3]{MootRetore2012lcg} for more details and references. 

\newcommand\type[1]{^{#1}}
\newcommand\et{\mathop{\&}}

\begin{figure}\label{church} 
$$
\begin{array}[t]{r|l} 
\multicolumn{2}{c}{\mbox{Logical\ connectives\ and\ quantifiers}}\\ 
\mbox{Constant} & \mbox{Type}\\ \hline 
	\textrm{\&, and} & \ttt \fl (\ttt \fl \ttt) \\ 
	\textrm{$\lor$, or} & \ttt \fl (\ttt \fl \ttt) \\ 
	\textrm{$\Rightarrow$, implies} & \ttt \fl (\ttt \fl \ttt)\\  \hline 
	\exists & (\eee \fl \ttt) \fl \ttt \\ 
	\forall & (\eee \fl \ttt) \fl \ttt \\ 
\end{array} 
$$
\caption{The logical constants and their types.}
\end{figure} 

A small example goes as follows. Assume that the  syntactical analysis of the sentence "\emph{Some club defeated Leeds.}" is 
\begin{center}
(some\ (club)) (defeated\ Leeds) 
\end{center} 
where the function is always the term on the left. If the semantic terms are as in the lexicon in figure \ref{semanticlexicon}, placing the semantical terms in place of the words yields a large $\lambda$-term that can be reduced:

\begin{figure}
\caption{A simple semantic lexicon} 
\begin{center} 
\begin{tabular}{ll} \hline 
\textbf{word} &  \textbf{\itshape semantic type $u^*$}\\ 
& \textbf{\itshape  semantics~: $\lambda$-term of type $u^*$}\\ 
&  {\itshape  $x\type{v}$ the variable or constant $x$ 
is of type $v$}\\ \hline 
\textit{some} 
& $(\eee\fl \ttt)\fl ((\eee\fl \ttt) \fl \ttt)$\\ 
& $\lambda P\type{\eee\fl \ttt}\  \lambda Q\type{\eee\fl \ttt}\  
(\exists\type{(\eee\fl \ttt)\fl \ttt}\  (\lambda x\type{\eee}  (\et\type{\ttt\fl (\ttt\fl \ttt)} (P\ x) (Q\ x))))$ \\  \hline 
\textit{club}  & $\eee\fl \ttt$\\ 
& $\lambda x\type{\eee} (\texttt{club}\type{\eee\fl \ttt}\  x)$\\  \hline 
\textit{defeated} & $\eee\fl (\eee \fl \ttt)$\\ 
& $\lambda y\type{\eee}\  \lambda x\type{\eee}\  ((\texttt{defeated}\type{\eee \fl (\eee \fl \ttt)}\  x)  y)$ \\  \hline 
\textit{Leeds} &$\eee$ \\ &  Leeds 
\end{tabular}
\end{center} 
\label{semanticlexicon}
\end{figure}

$$
\begin{array}{c} 
\Big(\big(\lambda P\type{\eee\fl \ttt}\ \lambda Q\type{\eee\fl \ttt}\  (\exists\type{(\eee\fl \ttt)\fl \ttt}\  (\lambda x\type{\eee}  (\et (P\ x) (Q\ x))))\big)
\big(\lambda x\type{\eee} (\texttt{club}\type{\eee\fl \ttt}\  x)\big)\Big) \\ 
\Big(
\big(\lambda y\type{\eee}\  \lambda x\type{\eee}\  ((\texttt{defeated}\type{\eee\fl (\eee\fl \ttt)}\  x)  y)\big)\ Leeds\type{\eee}\Big)\\ 
\multicolumn{1}{c}{\downarrow \beta}\\ 
\big(\lambda Q\type{\eee\fl \ttt}\  (\exists\type{(\eee\fl \ttt)\fl \ttt}\  (\lambda x\type{\eee}  (\et\type{\ttt\fl (\ttt\fl \ttt)}  
(\texttt{club}\type{\eee\fl \ttt}\  x) (Q\ x))))\big)\\ 
\big(\lambda x\type{\eee} \ ((\texttt{defeated}\type{\eee\fl (\eee \fl \ttt)}\  x)  Leeds\type{\eee})\big)\\ 
\multicolumn{1}{c}{\downarrow \beta}\\ 
\big(\exists\type{(\eee\fl \ttt)\fl \ttt}\  (\lambda x\type{\eee}  (\et (\texttt{club}\type{\eee\fl \ttt}\  x) ((\texttt{defeated}\type{\eee\fl (\eee\fl \ttt)}\  x)  Leeds\type{\eee})))\big)
\end{array}
$$ 

\noindent  which one usually writes as  $\exists x^\eee\quad club(x) \& defeated(x, Leeds^\eee)$

Clearly, it would be more accurate to have many individual base types rather than just $\eee$. Thus, the 
application of a predicate to an argument may only happen when it makes sense. 
For instance sentences like \ma{The chair barks.} or \ma{Their five is running.} are easily ruled out when there are several types for individuals 
by saying that \ma{barks} and \ma{is running}  
apply to individuals of type \ma{animal}. 
Nevertheless, such a type system needs to incorporate some flexibility. Indeed, in the context of a football match, the second sentence makes sense:  \ma{their five}  can be the player wearing the 5 shirt and who, being \ma{human}, is an \ma{animal} that can \ma{run}.

Our system is called the Montagovian Generative Lexicon or $\Lambda Ty_n$. Its lambda terms extend the simply typed ones of Montague semantics above. Indeed, we use second order lambda terms from Girard's   
system 
$\systF$ (1971) 
\citep{Girard2011blindspot}.

The types of \ltyn are defined as follows: 
 \begin{itemize} 
\item 
Constants types $\eee_i$ and $\ttt$, as well as type variables $\alpha,\beta,\ldots$ are types. 
\item 
$\Pi\alpha.\ T$ is a type whenever $T$ is a type and $\alpha$  a type variable . The type variable may or may not occur in the type $T$. 
\item 
$T_1\fl T_2$ is a type whenever $T_1$ and $T_2$ are types. 
\end{itemize}

The terms of $\ltyn$, are defined as follows: 
\begin{itemize} 
\item A variable  of type $T$ i.e. $x:T$ or  $x^{T}$  is a \emph{term}, and there are countably many variables of each type.
\item In each type, there can be a countable set of constants of this type, and a constant of type $T$ is a term of type $T$. Such constants are needed for logical operations and for the logical language (predicates, individuals, etc.). 
\item 
$(f\ t)$ is a term of type $U$ whenever $t$ is a term of type $T$ and  $f$ a term of type $T\fl U$. 
\item 
$\lambda x^{T}.\ \tau$ is a term of type $T\fl U$ whenever $x$ is variable of type $T$, 
and $t$ a term of type $U$.  
\item $t \{U\}$ is a term of type $T[U/\alpha]$
whenever $\tau$ is a term pf type $\Pi \alpha.\ T$, and $U$ is a type. 
\item $\Lambda \alpha. t$ is a term of type $\Pi \alpha. T$
whenever $\alpha$ is a type variable, and  $t:T$ a term without any free occurrence of the type variable $\alpha$ in the type of a free variable of $t$.  
\end{itemize}

The later restriction is the usual one on the proof rule for quantification in propositional logic: one should not conclude that $F[p]$ holds for any  proposition $p$
when assuming $G[p]$ --- i.e. having a free hypothesis of type $G[p]$. 

The reduction of the terms in system F or its specialised version \ltyn is defined by the two following reduction schemes that resemble each other:  
\begin{itemize} 
\item $(\lambda x. \tau) u$ reduces to $\tau[u/x]$ (usual $\beta$ reduction). 
\item $(\Lambda \alpha. \tau) \{U\}$  reduces to $\tau[U/\alpha]$ (remember that $\alpha$ and $U$ are types). 
\end{itemize} 

As  \citet{Girard71,Girard2011blindspot} showed 
reduction is strongly normalising and confluent 
\textit{every term of every type admits a unique normal form which is reached no matter how one proceeds.} 
This has  a good consequence for us, see e.g. \citep[Chapter 3]{MootRetore2012lcg}: 

\begin{quotation}\it 
\noindent \textbf{\ltyn\ terms as formulae of a many-sorted logic} 
If the predicates, the constants and the logical connectives and quantifiers 
are the ones  from a many sorted logic of order $n$ (possibly $n=\omega$) then the closed normal terms of $\ltyn$ of type $\ttt$ unambiguously correspond to many sorted  formulae of order $n$. 
\end{quotation} 

The polymorphism of system $\systF$ is a welcome simplification.  
For instance, a single constant $\exists$ of type 
$\Pi \alpha. (\alpha \fl \ttt) \fl \ttt$ is enough for the family of existential quantifiers over all possible types. Indeed such a polymorphic type  can be specialised to a specific type $T$, yielding the properly typed existential quantifier over $T$. 

\begin{figure} 
\label{polyandfig} 
\begin{center}
\includegraphics[scale=0.2]{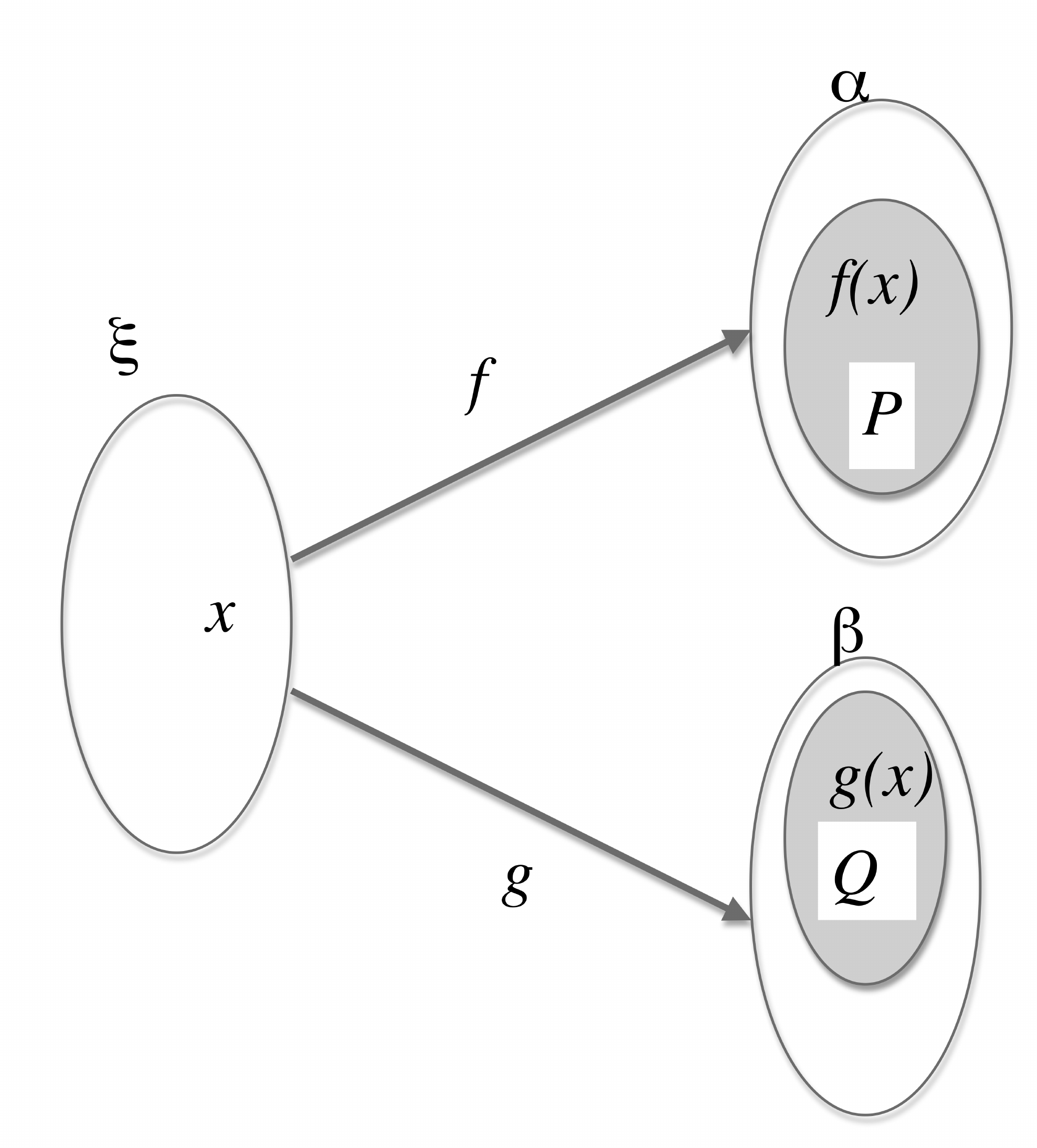} 
\end{center} 
\caption{\large Polymorphic and: $P(f(x))\& Q(g(x))$  with $x:\xi$, $f:\xi\fl\alpha$, $g:\xi\fl\beta$.}  
\end{figure}

Polymorphism also allows a factored treatment of conjunction for copredication: \emph{whenever} 
an object $x$ of type $\xi$ can be viewed both as an object 
of type $\alpha$ to which a property $P^{\alpha\fl\ttt}$ applies and as an object of type $\beta$  
to which a property $Q^{\beta\fl\ttt}$ applies
(via two terms $f_0:\xi\fl\alpha$ and $g_0:\xi\fl\beta$ ), the fact that 
$x$ enjoys  $P\land Q$ can be expressed by the unique polymorphic term (see explanation in figure \ref{polyandfig}): 
\begin{exe} 
\ex \label{polyandterm} 
$\Land=\Lambda \alpha \Lambda \beta
\lambda P^{\alpha \fl \ttt} \lambda Q^{\beta\fl \ttt} 
 \Lambda \xi \lambda x^\xi 
 \lambda f^{\xi\fl\alpha} \lambda g^{\xi\fl\beta}.\\ 
\hspace*{15em}\hfill (\et^{\ttt\fl\ttt\fl\ttt} \ (P \ (f \ x)) (Q \ (g \  x))) 
$
\end{exe} 

The lexicon provides each word with:

\begin{itemize} 
\item 
A main $\lambda$-term of \ltyn, the``usual one" specifying the argument structure of the word.
 \item 
A finite number of $\lambda$-terms of \ltyn\ (possibly none) that implement meaning transfers. Each of this meaning transfer is declared in the lexicon to be \emph{flexible} \flex\ or \emph{rigid} \rig. 
\end{itemize}

\begin{figure} 
\caption{A sample lexicon}
$$ 
\begin{array}{l|l|rl} 
\mbox{word} & \mbox{principal\ $\lambda$-term} & \multicolumn{1}{l}{\mbox{optional\ $\lambda$-terms}} & \mbox{rigid/flexible}\\ \hline 
Liverpool & \lpl^T & Id_T:T\fl T &\flex \\ 
& & t_1:T\fl F &\rig \\ 
& & t_2:T\fl P &\flex \\ 
& & t_3:T\fl Pl &\flex\\ 
\hline 
spread\_out & spread\_out:Pl\fl\ttt & \\ 
\hline 
voted & voted:P\fl\ttt& \\
\hline 
won & won:F\fl\ttt&\\  
\end{array}
$$
where the base types are defined as follows: 
\begin{tabular}[t]{ll}
$T$ & town \\ 
$P$ & people \\ 
$Pl$ & place \\
\end{tabular} 
\label{lexicon} 
\end{figure}

Let us see how such a lexicon works. 
When a predication requires a type $\psi$ (e.g. Place) while its argument is of type $\sigma$ (e.g. Town)
the optional terms in the lexicon can be used to ``convert" a Town into a Place.

\begin{exe}
\ex \label{lplex}
\begin{xlist} 
\ex Liverpool is spread out. \label{ll} 
\ex 
This sentence leads to a type mismatch $spread\_out^{Pl\fl\ttt}(\lpl^T))$, since \ma{spread\_out} applies to \ma{places} (type $Pl$) and not to \ma{towns} as \ma{Liverpool}. 
This type conflict  is solved using the optional term $t_3^{T\fl Pl}$ provided by the entry for \ma{Liverpool}, which turns a town ($T$) into a place ($Pl$) 
\linebreak $spread\_out^{Pl\fl\ttt}(t_3^{T\fl Pl} \lpl^T))$ --- a single optional term is used, the \flex / \rig difference is useless. 
\end{xlist} 
\ex
\begin{xlist}
\ex 
Liverpool is spread\_out and voted (last Sunday).  \label{llv} 
\ex 
In this example, the fact that \ma{Liverpool} is \ma{spread\_out} is derived as previously, and the fact \ma{Liverpool} \ma{voted} is obtained from the transformation of the town into people, which can vote. The two can be conjoined by the polymorphic \ma{and} defined above  in \ref{polyandterm} ($\Land$) 
because these transformations are flexible: one can use both of them.  
We can make this precise using only the rules of our typed calculus. 
The syntax yields the predicate $(\Land (is\_spread\_out)^{Pl\fl \ttt} (voted)^{P\fl\ttt})$ and consequently 
the type variables should be instantiated by $\alpha:=Pl$ and $\beta:=P$ and the exact term is 
$\Land \{Pl\} \{P\} (is\_spread\_out)^{Pl\fl \ttt} (voted)^{P\fl\ttt}$ which reduces to:  

$ \Lambda \xi \lambda x^\xi  \ 
 \lambda f^{\xi\fl\alpha} \lambda g^{\xi\fl\beta}  
(\et^{\ttt\fl\ttt)\fl\ttt} \ (is\_spread\_out \ (f \ x)) (voted \ (g \  x)))$. 

Syntax also says this term is applied to \ma{Liverpool}. 
which forces the instantiation $\xi:=T$ and the term corresponding to the sentence is after some reduction steps,\\  
$ \lambda f^{T\fl Pl} \lambda g^{T\fl P}  
(\et \ (is\_spread\_out \ (f \ \lpl^T)) (voted \ (g \  \lpl^T))))$. Fortunately the optional $\lambda$-terms 
$t_2:T\fl P$ and  $t_3:T\fl Pl$ are provided by the lexicon, and they can both be used, since none of them is rigid.
Thus we obtain, as expected\\  
$(\et \ (is\_spread\_out^{Pl\fl\ttt} \ (t_3^{T\fl Pl} \ \lpl^T)) (voted^{Pl\fl\ttt} \ (t_2^{T\fl P} \  \lpl^T)))$ 
\end{xlist}
\ex 
\begin{xlist} 
\ex \label{lvw}    \#
Liverpool voted and won (last Sunday). 
\ex 
This third and last example is rejected as expected. Indeed, the transformation of the town into a football club prevents any other transformation (even the identity) to be used with 
the polymorphic \ma{and} ($\Land$) defined above  in \ref{polyandterm}. 
We obtain the same term as above, with $won$ instead of $is\_spread\_out$:  $ \lambda f^{T\fl Pl} \lambda g^{T\fl P}  
(\et \ (won \ (f \ \lpl^T)) (voted \ (g \  \lpl^T))))$ 
and the lexicon provides the two morphisms that would solve the type conflict, but one of them is 
\emph{rigid}, i.e. we can solely use  this one. Consequently no semantics can be derived from this sentence,  which is semantically invalid. 
\end{xlist} 
\end{exe}

%
%


The difference between our system end those of  \citet{Luo2011lacl,asher-webofwords}. 
does not rely on the type systems, which are quite similar, 
but in the \emph{architecture} which  is, in our case,  rather \emph{word driven} than type driven. The optional morphisms are anchored in the words, and do not derive from the types. This is supported in our opinion by the fact that some words with the very same ontological type (like French nouns  \ma{classe} and \ma{promotion}, that are groups of students in the context of teaching) may undergo different coercions (only the first one can mean a classroom). 
This rather lexicalist view goes well with the present work that proposes to have specific entries for deverbals, 
that are derived from the verb entry but not automatically. 

This system has been implemented as an extension to the Grail parser \citet{moot10grail}, with $\lambda$-DRT instead of formulae as $\lambda$-terms. 
It works fine once the semantic lexicon has been typeset.\footnote{Syntactical  categories are learnt from annotated corpora, but  semantical typed $\lambda$-terms cannot yet, as discussed in the conclusion.}

We already explored some of  the compositional properties (quantifiers, plurals and generic elements,....) of our Montagovian generative lexicon as well as some 
the lexical issues (meaning transfers, copredication, fictive motion,...  )  
\citep{BMRjolli,Retore2012rlv,Retore2013taln,MMR2013lenls}. 

\section{Deverbals in the Montagovian generative lexicon} 

\subsection{Deverbals senses in the Montagovian generative lexicon} 

In our richly typed view of Montague semantics, a verb is a many-sorted  predicate, whose arguments (i.e. complements, subject, adjuncts, etc.) are properly typed.  As established above, one cannot predict what aspect(s) of the verb meaning will be selected by a given nominalisation suffix, hence the possible meanings have to be specified in the lexicon. 

Hence the deverbal requires a specific polysemous entry with  optional morphisms. 
As the principal sense, we chose the process itself , as expressed by the base verb,
\footnote{If there were entries for abstract event we would chose them rather than
the base verb, but we have no dictionary of events while we do have precise electronic dictionaries, e.g. the TLFi \citep{Pierrel2006tlfi}) which include the verbs.} of type $\vvv$ 
when this sense is available,
and any other sense otherwise. 
As the case of \emph{garer/garage} 
shows,  a relevant meaning need not be an actant (complement, subject) but can be an adjunct (that is an unnecessary complement, the locative complement in the case of \emph{garage} ). In the formal organisation of an entry, it is fairly natural to consider the processual meaning as the main sense, when it exists, since all other possible senses have to do with the process.  Observe that this meaning may not exist anymore: in contemporary French, the \emph{garage} is no longer understood as a process. 
Optional morphisms turns the process of the verb of type $\vvv$ (or the main meaning of type of the deverbal $\xxx$ if it is not $\vvv$) into other types corresponding to the other senses.  

For instance, \ma{assinatura} (Brazilian Portuguese) enjoys three readings:  
\begin{itemize}
\item the whole process yielding to a signature (discussions leading to an agreement and concluded by a writing act), 
\item the writing act itself,
\item the grapheme that results from the writing act. 
\end{itemize} 

Hence the lexical entry associated with \ma{assinatura} 
contains two optional morphisms, one $f_\phi$ of type $\vvv\fl\vvv_\phi$ 
which turn the event into a physical event  of type $\vvv_\phi$  (specialised events that modify the physical world) and   $f_\phi$ of type $\vvv\fl\phi$ that turns the event into a physical object of type $\phi$,
as can be seen in figure \ref{lexicon}. 

\begin{exe}
\ex \label{assinaturax3}
\begin{xlist}
\ex \label{atrasou} A assinatura atrasou tr\^es dias.\footnote{``The signing was delayed by three days.'' Example from \url{http://noticias.uol.com.br/inter/efe/2004/03/05/ult1808u6970.jhtm}.} 
\ex \label{ilegivel} A assinatura estava ileg\'\i vel.\footnote{``The signature was illegible.'' Example from \url{http://www.reclameaqui.com.br/3372739/dix-saude/cancelamento-do-plano-a-mais-de-um-mes-e-nada/}.}
\ex \label{furou} A assinatura furou a folha.\footnote{Home made example tested on native speakers.}
\end{xlist} 
\end{exe} 


\subsection{Felicitous and infelicitous copredications on different aspects of a deverbal} 

Usually copredication between process and any other aspect sounds weird, for instance one cannot conjoin the properties in example \ref{atrasouelegivel} --- but it is possible that some senses are  compatible, like \ref{furouelegivel}. 

\begin{exe} 
\ex[*] {A assinatura atrasou tr\^e s dias e estava ileg\'\i vel.} \label{atrasouelegivel} 
\trans `the signature took three days and is illegible' 
\ex \label{furouelegivel} A assinatura furou a folha e estava eleg\'\i vel.  
\trans `the signature pierced the sheet of paper and is illegible.' 
\end{exe} 

How does the Montagovian generative lexicon accounts for possible and impossible copredications on deverbals? Deverbal are handled just as other polysemous nouns, that is as  explained in examples \ref{lplex}, using the \emph{rigid/flexible distinction} and 
the polymorphic \ma{and} $\Land$  defined in   (\ref{polyandterm}). It should be observed that for \ma{assinatura} the identity is declared to be rigid, thus expressing that the event sense is  incompatible with other senses.

\begin{figure} 
\caption{The lexical entry for \ma{assinatura}}
$$ 
\begin{array}{l|l|rl} 
\mbox{word} & \mbox{principal\ $\lambda$-term} & \multicolumn{1}{l}{\mbox{optional\ $\lambda$-terms}} & \mbox{rigid/flexible}\\ \hline 
assinatura & \lambda x^\vvv.\ (\assi(x)):\vvv\fl\ttt & Id_\vvv:\vvv\fl\vvv &\rig \\ 
&& f_{\vvv_\phi}:{\vvv\fl\vvv_\phi}  & \flex \\ 
&& f_\phi:{\vvv\fl\phi} & \flex \\ 
\end{array}
$$
where the base types are defined as follows: 
\begin{tabular}[t]{ll}
$\vvv$ & events \\ 
$\phi$ & physical objects \\ 
$\vvv_\phi$ & physical events \\
\end{tabular} 
\label{entryassinatura} 
\end{figure}

The lexical entry for ``assinatura"  (\emph{signature}, whose type is $\vvv$) is given in figure \ref{entryassinatura}:
\begin{exe}
\ex 
the main term is $\lambda x^{\vvv}.(\assi^{\vvv \fl t} x)$   
\ex 
and the optional morphisms are 
\begin{xlist}
\ex 
$Id = \lambda x^{\vvv} . x$, the (always present) identity (referring to the agreement process)  which is declared as \emph{rigid}   
\ex 
 $f_{\phi_\phi}^{\vvv\fl \vvv_\phi}$ turning the event into a physical event and declared to be flexible 
\ex 
$f_{\phi}^{\vvv \rightarrow \phi}$ turning the event into a physical object and declared to be flexible 
 \end{xlist}
 \end{exe} 
 
A first step is the composition of \ma{assinatura} with the definite article ``a'' (``the"). 
As explained in  \citep{Retore2013taln} where more details can be found, 
the definite determiner is handled by a typed choice function:  
$$\iota:\Lambda \alpha.(\alpha \rightarrow \ttt) \rightarrow \alpha$$ 
 When this polymorphic  $\iota$ ($\Lambda \alpha...$) is specialised to the type $\vvv$ ($\alpha:=\vvv$)
 and applied to the predicate $\assi:(\vvv\rightarrow \ttt)$  it yields 
 $\iota \{\vvv\} \assi$ of type $\vvv$ whose  short hand in the examples is written $(\sig)^{\vvv}$. 
 This term introduces a presupposition: $\assi(\iota(\assi))$, saying that the designed event is an \ma{assinatura}.

Here are some more shorthands for handling the examples:
\begin{itemize} 
\item  $\atratres:(v \rightarrow t)$ stands for the predicate \ma{atrasou tr\^e s dias}  (\textit{took three days}) which applies to events 
\item $\ilg:\phi\rightarrow t$ stands for  the  predicate \ma{estava ileg\'\i vel} (\textit{was illegible}) that applies to physical objects; 
\item  $\furou:\vvv_\phi\fl\ttt$ stands for the predicate \ma{a furou a folha} which applies to an event which affects the material world. 
\end{itemize}
The semantics terms of these complex predicates are computed from the entries in the lexicon, but as this is just standard Montague semantics, we leave out the details. 
 
%
%
%
%

Let us recall the polymorphic \ma{and} from (\ref{polyandterm}): 

$\Land=\Lambda \alpha \Lambda \beta
\lambda P^{\alpha \fl \ttt} \lambda Q^{\beta\fl \ttt} 
 \Lambda \xi \lambda x^\xi 
 \lambda f^{\xi\fl\alpha} \lambda g^{\xi\fl\beta}.\\ 
\hspace*{15em}\hfill (\et^{\ttt\fl\ttt\fl\ttt} \ (P \ (f \ x)) (Q \ (g \  x))) 
$

The instantiations for our example should be as follows: 
\begin{itemize} 
\item 
$\alpha= \vvv_\phi$, $P= \furou^{\vvv_\phi\fl\ttt}$, $f=f_{\vvv_\phi}$, 
\item 
$\beta=\phi$, $Q=\ilg^{\phi\fl\ttt}$,  $g=f_{\phi}$, 
\item 
$ \xi=\vvv$, $x=\sig^{\vvv}$. 
\end{itemize} 
The polymorphic 
``and" $\Land$ takes as arguments two properties $P$ (here: $\furou$ ) and Q (here: $\ilg$)
of  entities of respective type $\alpha$ (here: $\vvv_\phi$) and $\beta$ (here: $\phi$), and returns a predicate that applies to a term $x$ of type $\xi$. This predicate says that 
\begin{itemize}
\item 
if $x$ of type $\xi$ (here $\sig$ of type $\vvv$): 
\begin{itemize} 
\item 
enjoys $P$ (here $\furou(f_{\vvv_\phi}(x))$)
when  viewed as an object of type $\alpha$ (here $\vvv$) via  $f_{\vvv_\phi}$  
\item 
enjoys $Q$ (here $\ilg(f_\phi(x))$) 
when   viewed as an object of type $\beta$ (here $\phi$)  via some $g$ (here $f_\phi$) 
\end{itemize} 
\item 
then $x$ endowed with the proper meanings has both properties that is  $$\furou(f_{\vvv_\phi}(x)) \& \ilg(f_\phi(x))$$
\end{itemize} 
Hence the copredication in example (\ref{furouelegivel}) can be derived.

The co-predication in example (\ref{atrasouelegivel})  involving the predicates  ``took three days" ($\atratres$) and ``was illegible" ($\ilg$) works just the same but to conjoin 
the predicates with the polymorphic $\Land$  the instantiation should be: 
\begin{itemize} 
\item 
$P= \atratres$, $\alpha= \vvv$, $f=Id_{\vvv}$, 
\item 
$Q=\ilg$, $\beta=\phi$, $g=f_{\phi}$, 
\item 
$ \xi=\vvv$, $x=\sig^{\vvv}$. 
\end{itemize} 
Thus we both use $f=Id_{\vvv}$ and $g=f_{\phi}$ which is impossible, because $f=Id_{\vvv}$ (referring to the process itself, from discussion to agreement and signature) is declared to be rigid in the lexicon.


\section{Conclusions: limits and future works}

The Montagovian Generative Lexicon with word-driven meaning-transfers offers a natural way to interpret the polysemy of deverbals. This model also handles a treatment of the (in)felicity  of copredications, 
generating the felicitous ones and blocking the infelicitous ones.  
This work has been implemented as an extension of the Grail syntactic and semantic parser with small hand typed semantic lexicons, in particular for motion verbs and the corresponding deverbals (\emph{arriver/arriv\'ee, chemin/cheminer, partir/d\'epart}) for the study of an historical and regional corpus of travel stories, see  \citep{moot10grail,LMRS2012taln}. 

As we have seen, the semantics of verbal nominalisations  is only partly inferable from the event verb. 
The possibles senses are derivable from the event, since they are its subject, one of its complement or one of its adjuncts, but the one that are selected cannot be predicted from the suffix --- sometimes the event itself is not (anymore) a possible sense. 
Furthermore the felicity of copredications on different sense is rather unpredictable. 

So our formalisation suffers from this lack of general rules which could produce the deverbal entries from the verb entries: one has to define each lexical entry patiently  for defining an automated semantic analysis rejecting illformed sentences. 
As suggested by reviewers there are two methods that could find regularities in the senses of verbal nominalisations, and in the felicity of copredications: 
\begin{itemize}
\item a diachronic study of the evolution of deverbal meanings: how did the deverbal meaning evolve? is the event sense always  the initial one? 
\item the study of (first) language acquisition: how do we learn the possible meanings of a deverbal? 
\end{itemize} 
On the practical side, i.e. for the automatic acquisition of such a semantic lexicon, some rules or even the entries themselves could be inferred by using machine learning and distributional semantics, which is by now highly efficient to infer rules between words and weighted semantical relations. Such a learning task is very appealing to us. \citep{VandeCruys2010Mining,LafourcadeJoubert2010imcsit,Lafourcade2011hdr}

Besides, there are two points that seem rather easy to improve. 

Firstly, we restricted impossible copredications to the case where one meaning is incompatible with any other meaning.  One could think of more complicated incompatibilities, e.g. of three senses that are pairwise compatible but the three of them being incompatible. We are presently looking for such examples e.g. with towns that have a rich polysemy, and we developed in \citet{MR2013nlcs} a modelling of these complicated constraints 
using linear logic --- see \citet{Girard2011blindspot} for a recent presentation. 

Secondly, we can say something about the relation to pragmatics and the influence of the context. 
 The examples below, in German and from \cite[p.170]{bra11} show as we cannot predict the behaviour of AdjectiveNouns compounds in contexts, without a contextual or idyosincratic component:
 \begin{exe} 
 \ex 
 \begin{xlist} 
\ex Die langwierige Übersetzung brachte mir viel Geld ein.
\ex 
`The tedious translation earned me a lot of money.’ 
\end{xlist} 
\ex 
\begin{xlist} 
\ex ?Die einfache Übersetzung brachte mir viel Geld ein. 
\footnote{Observe that with an opposition the sentence would be fine: 
\ma{This translation although quite easy, earned me a lot of money.}} 
\ex 
`The easy translation earned me a lot of money.’ 
\end{xlist} 
\ex 
\begin{xlist} 
\ex Die einfache Übersetzung brachte mir dennoch viel Geld ein. 
\ex 
`The easy translation still earned me a lot of money.’
\end{xlist} 
\end{exe}
As we showed before, the same happens, at least, in Portuguese (\ref{fritura} and \ref{ainda}), English (\ref{tn} and \ref{tn1}) and French (\ref{bar} and \ref{bar1}).
Such an infringement of the usual lexical rules for felicity rely on the context. Once more some cases could possibly be solved by distributional semantics and lexical networks \citep{VandeCruys2010Mining,LafourcadeJoubert2010imcsit,Lafourcade2011hdr}: in particular if the predicates are similar enough, the usually infelicitous copredications may become felicitous.  As similarity measures do exists for word and for complex predicates as well, we can possibly incorporate in our model these pragmatic phenomena 
which derive from the context.

\paragraph{Thanks:} We are indebted to Christian Bassac (University of Lyons 2), Marcos Lopes
(University of S\~ao Paulo),
Urszula Wybraniec-Skardowska (University of Opole), and to the anonymous referees for their helpful comments.

\end{document}